\documentclass[]{spie}  

\usepackage{amsmath,amsfonts,amssymb}
\usepackage{graphicx}
\usepackage[colorlinks=true, allcolors=blue]{hyperref}
\usepackage{subcaption}
\usepackage[labelfont=bf, width=.8\textwidth]{caption}
\usepackage{array} 
\usepackage{geometry}
\usepackage{tabu}                     
\usepackage{multirow}                 
\usepackage{multicol}                 
\usepackage{multirow}                
\usepackage{float}                    
\usepackage{makecell}                 
\usepackage{booktabs}                 
\usepackage{algorithm}
\usepackage{algorithmic}
\usepackage{lipsum} 
\usepackage{hyperref}
\usepackage{cleveref}
\usepackage{adjustbox}

\title{Towards Fine-grained Renal Vasculature Segmentation: Full-Scale Hierarchical Learning with FH-Seg}

\author[a]{Yitian Long}
\author[b]{Zhongze Wu}
\author[b]{Xiu Su}
\author[c]{Lining Yu}
\author[c]{Ruining Deng}
\author[d]{Haichun Yang}
\author[a,c,d]{Yuankai Huo}

\affil[a]{Data Science Institute, Vanderbilt University, Nashville, TN, USA}
\affil[b]{Big Data Institute, Central South University, Changsha, Hunan, China}
\affil[c]{Department of Computer Science, Vanderbilt University, Nashville, TN, USA}
\affil[d]{Department of Pathology, Microbiology and Immunology, Vanderbilt University Medical Center, Nashville, TN, USA}

\authorinfo{First author: Yitian Long: E-mail: yitian.long@vanderbilt.edu
\\Corresponding author: Yuankai Huo: E-mail: yuankai.huo@vanderbilt.edu\\}

\begin{document} 
\maketitle
\vskip -0.2in
\begin{abstract}
Accurate fine-grained segmentation of the renal vasculature is critical for nephrological analysis, yet it faces challenges due to diverse and insufficiently annotated images. Existing methods struggle to accurately segment intricate regions of the renal vasculature, such as the inner and outer walls, arteries and lesions. In this paper, we introduce FH-Seg, a Full-scale Hierarchical Learning Framework designed for comprehensive segmentation of the renal vasculature. Specifically, FH-Seg employs full-scale skip connections that merge detailed anatomical information with contextual semantics across scales, effectively bridging the gap between structural and pathological contexts. Additionally, we implement a learnable hierarchical soft attention gates to adaptively reduce interference from non-core information, enhancing the focus on critical vascular features. To advance research on renal pathology segmentation, we also developed a Large Renal Vasculature (LRV) dataset, which contains 16,212 fine-grained annotated images of 5,600 renal arteries. Extensive experiments on the LRV dataset demonstrate FH-Seg’s superior accuracies (71.23\% Dice, 73.06\% F1), outperforming Omni-Seg by 2.67 and 2.13 percentage points respectively. Code is available at: https://github.com/hrlblab/FH-seg.
\end{abstract}

\section{INTRODUCTION}
\label{sec:intro}  

 Pathological image segmentation techniques \cite{gomes2021building,marti2021digital} play a crucial role in renal pathology. Advances in pixel-level segmentation models, such as U-Net \cite{ronneberger2015u,oktay2018attention,huang2020unet,jayapandian2021development}, DeepLabV3 \cite{lutnick2019integrated}, and Swin Unetr \cite{hatamizadeh2021swin}, have significantly improved medical image segmentation. However, fine-grained segmentation of the renal vasculature, including key structures such as inner and outer walls, arteries, and small lesions, remains challenging \cite{he2020dense,lutnick2021mo077,Wang2022An}, especially in settings with limited annotation \cite{he2021meta,xu2023hybrid}. Digital pathology has transformed renal disease diagnosis and prognosis by integrating artificial intelligence to enhance diagnostic accuracy and efficiency. A detailed segmentation of vascular changes is essential for understanding renal disease processes \cite{mounier2002cortical,bellomo2012acute,jimenez2006mast}.

Manual segmentation of renal blood vessels is labor-intensive and prone to error, and existing methods still struggle to address the challenges of semantic segmentation and quantification due to the complex interrelationships between tissue structures. Recent studies have explored multi-scale spatial awareness techniques \cite{Omni-seg} to address these challenges. Matos et al. \cite{matos2024cpp} integrated pyramid pooling and dilated spatial pyramid pooling into U-Net, enhancing the capture of multi-scale contextual information and improving kidney tumor segmentation accuracy, but it may lose high-resolution features in small lesions. Cao et al. \cite{cao2024rasnet} proposed a multi-scale spatial perception module to ensure the integrity of kidney segmentation, but this increases network complexity. Cai et al. \cite{cai2022ma} introduced attention gates (AG) and multi-scale fusion to enhance the capture of both local and global information in kidney segmentation, but still faces the issue of detail loss. However, current methods face two primary limitations \cite{salvi2020karpinski,huo2021ai}: (1) difficulty in managing complex spatial relationships and scale variations among fine-grained vasculature types, and (2) interference from noise due to incomplete tissue segmentation.

    
    
    

\newpage
\begin{figure}[H]
    \centering
    \begin{minipage}[t]{0.55\textwidth}
        \vspace*{0pt}
        \includegraphics[width=\textwidth]{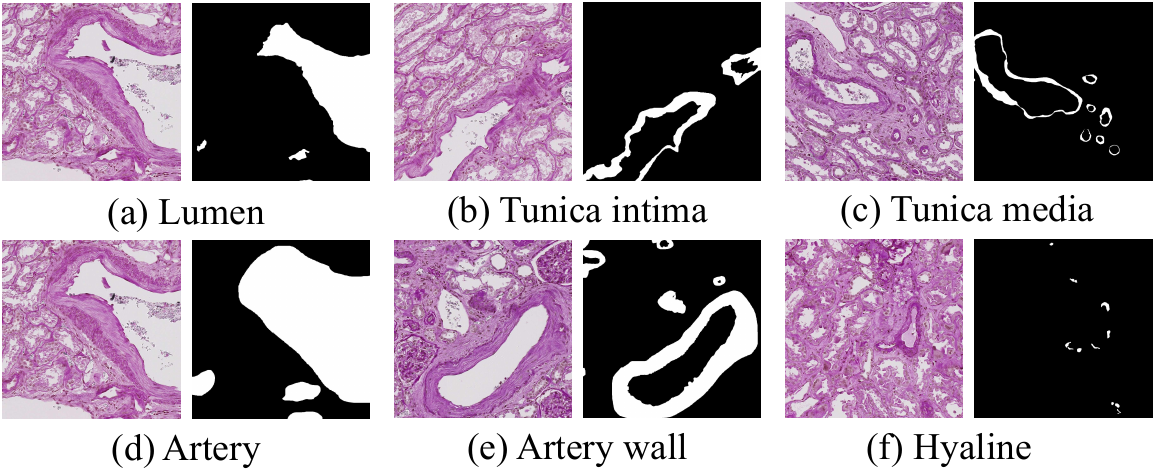}
        \caption{Examples of image-mask pairs of LRV dataset.}
        \label{FIG:LRV}
    \end{minipage}%
    \hfill
    \begin{minipage}[t]{0.4\textwidth}
        \vspace*{0pt}
        \captionof{table}{Summary of the dataset.}
        \label{tab:dataset_partition_summary}
        \setlength{\tabcolsep}{2pt} 
        \renewcommand{\arraystretch}{1} 
        \resizebox{\textwidth}{!}{
            \begin{tabular}{l|cccccc}
            \toprule
            \textbf{Category} & \textbf{Train} & \textbf{Val} & \textbf{Test} & \textbf{Total} & \textbf{Size} \\ 
            \midrule
            Lumen        & 1781 & 593 & 595 & 2969 & $2048^2$ \\ 
            Tunica intima    & 1694 & 564 & 566 & 2824 & $2048^2$ \\ 
            Tunica media     & 1932 & 644 & 644 & 3220 & $2048^2$ \\ 
            Artery       & 1932 & 644 & 645 & 3221 & $2048^2$ \\ 
            Artery wall      & 1932 & 644 & 644 & 3220 & $2048^2$ \\ 
            Hyaline       & 454  & 151 & 153 & 758  & $2048^2$ \\ 
            \midrule
            \textbf{Total} & 9725 & 3240 & 3247 & 16212 & - \\ 
            \bottomrule
            \end{tabular}
        }
    \label{tab:dataset}
    \end{minipage}
\end{figure}


In this paper, we introduce the \textbf{F}ull-Scale \textbf{H}ierarchical Learning Sgementation Framework (FH-Seg) for comprehensive segmentation of the renal vasculature as shown in \cref{FIG:1}. We first construct a detailed annotated renal vascular dataset, containing about 5600 arteries, including various complex structures of the renal vasculature, as shown in \cref{FIG:LRV} and \cref{tab:dataset}. Additionally, we develop a Full-scale Skip (FS) strategy to capture fine-grained details and coarse-grained semantic information across scales. Furthermore, we introduce a Hierarchical Soft Attention Gate (HSA) mechanism strategically integrated after the full-scale skip connections and before each upsampling step, adaptively reduce the interference of non-core information. The proposed method demonstrates improved performance on the LRV dataset, exceeding Omni-seg \cite{Omni-seg} by 2.67\% and 2.13\% in Dice and F1 accuracy, respectively.

\section{Related Work}
\subsection{Renal Pathology Segmentation}
\label{sec:title}
With the rapid advancements in deep learning, convolutional neural networks (CNNs) have become the standard approach for image segmentation \cite{feng2023artificial,hara2022evaluating}. Bueno et al. \cite{bueno2020glomerulosclerosis} proposed SegNet-VGG 16 to detect glomerular structures through multi-class learning, achieving a high Dice Similarity Coefficient (DSC). Lutnick et al. \cite{lutnick2019integrated} used DeepLab v2 to detect glomerulosclerosis, interstitial fibrosis, and tubular atrophy. Salvi et al. \cite{salvi2021automated} designed a network based on a multi-residual U-Net for quantitative analysis of glomeruli and renal tubules. Bouteldja et al. \cite{bouteldja2021deep} developed a CNN model capable of automatic multi-class segmentation of renal pathology across different mammalian species and experimental disease models. However, most of these methods still face challenges in the fine segmentation of renal vasculature, particularly under limited annotation conditions, including key structures such as inner and outer vessel walls, arteries, and small lesions.

\subsection{Attention Modules}
\label{sec:title}
Unlike natural images, pathological images contain complex cellular structures, dense tissue backgrounds, and significant noise. Attention modules can automatically adjust weights across different scales of feature maps, effectively integrating global and local information, ensuring that the model performs well in segmenting both small lesions and large structures \cite{azad2022smu}. However, modeling the complex spatial relationships and detailed structures in pathological images for segmentation models remains challenging. Several deep learning-based methods have been developed to capture spatial dependencies between pixels in feature maps \cite{fan2020ma,guo2021sa,azad2021deep}. These studies introduced attention gates and other mechanisms but primarily focused on general features in natural images. In contrast, our method applies a hierarchical soft attention mechanism explicitly designed for pathological contexts, enhancing core vascular feature segmentation by adaptively reducing noise interference.

\subsection{Multi-scale Segmentation}
\label{sec:title}
In renal pathological image segmentation, significant variation in scale among different objects is a key challenge. Deng et al. \cite{Deng2023omni} proposed a scale-aware dynamic network, which employs multiple segmentation networks or multi-head structures to address these diversities. However, such strategies often increase model complexity and lack explicit modeling of spatial relationships, limiting performance. Recent efforts have shown that combining multi-scale skip connections and attention modules can address these issues. For example, U-Net++ \cite{zhou2018unet++} introduced nested skip connections, and Attention U-Net \cite{oktay2018attention} used attention gates, but these studies did not investigate their combined use. To the best of our knowledge, our work is the first to integrate full-scale skip connections with hierarchical soft attention gates in a unified framework for renal vasculature segmentation.

In our proposed network, we explicitly designed the architecture with full-scale skip connections, integrating detailed anatomical information and cross-scale contextual semantics. We proposed a full-scale hierarchical learning framework (FH-Seg) that effectively bridges the gap between structural and pathological contexts. Additionally, we implemented a learnable hierarchical soft attention gate that adaptively reduces interference from non-core information, enhancing the focus on critical vascular features.




\begin{figure}       
	\centering
    \includegraphics[scale=0.5]{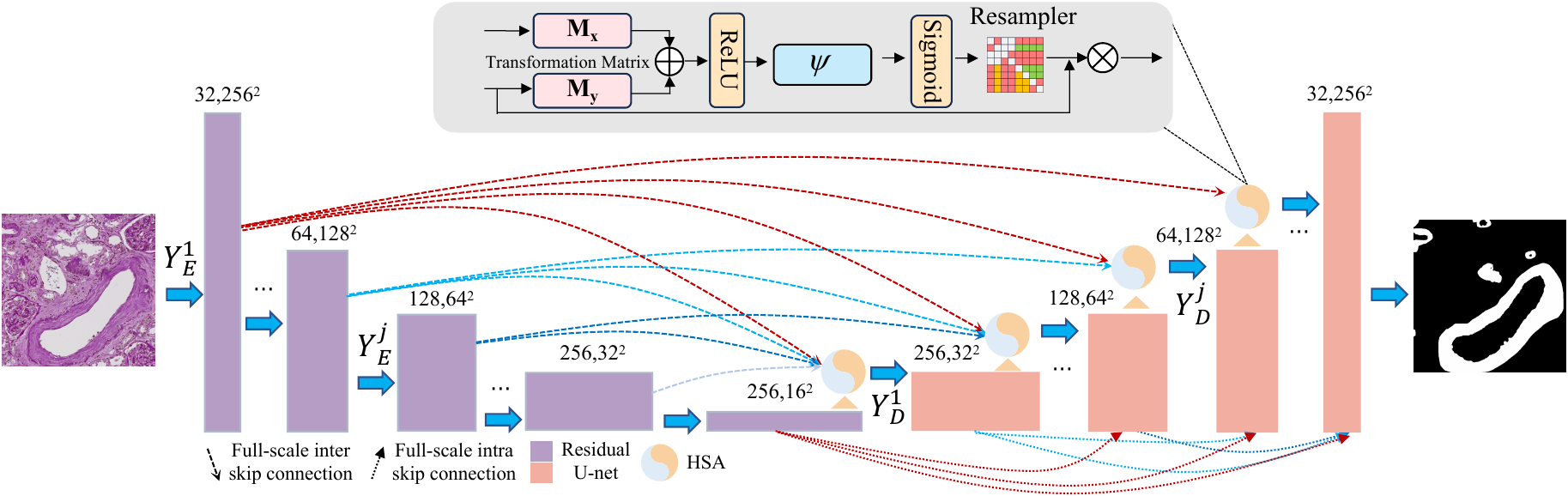}  
	\caption{Full-Scale Hierarchical Learning Segmentation Framework.}   
	\label{FIG:1}
 \vskip -0.2in
\end{figure}


\section{Methods}
\label{method}


\subsection{Preliminaries}
The traditional U-Net employs symmetric skip connections that allow each decoder level to directly receive input from its corresponding encoder level:
\begin{equation}
X_{De}^i = \mathcal{U}(X_{De}^{i+1}) \oplus X_{En}^i,
\end{equation}
where $X_{De}^i$ and $X_{En}^i$ represents the feature map at the $i$-th level of the decoder and encoder repectively, $\mathcal{U}$ denotes the upsampling operation, and $\oplus$ signifies the concatenation of feature maps.

The efficiency of U-Net in terms of parameter usage is notable. The encoder and decoder are symmetrical, with each layer $i$ having $32 \times 2^i$ channels. 
\begin{equation}
N_{De}^i = K_s \times K_s \times \left[ c(X_{D e}^{i+1}) \times c(X_{D e}^i) + c(X_{D e}^i)^2 + c(X_{En}^i + X_{D e}^i) \times c(X_{D e}^i) \right],
\end{equation}
where $K_s$ is the convolution kernel size, and $c(\cdot)$ denotes the channel depth.

\subsection{Model Architecture}
The FH-Seg framework employs a Residual U-Net backbone with full-scale skip connections $\mathcal{S}_F$ and hierarchical soft-attention gates $\mathcal{A}_{\text{HSA}}$, depicted in \cref{FIG:1}. The model combines detailed anatomical information with contextual semantics across scales to enhance fine-grained segmentation performance. The inclusion of the ``Resampler" in $\mathcal{A}_{\text{HSA}}$ ensures feature maps are normalized to the same spatial resolution before being weighted, thereby improving compatibility between multi-scale features.

\subsection{Full-scale Skip Connections}
The renal vasculature structures present complex segmentation challenges due to their subtle tissue variations and intricate voids, as shown in \cref{FIG:LRV}. To address this, our FH-Seg model introduces full-scale skip connections $\mathcal{S}_F$ to integrate multi-level semantic and detailed features across different scales. 

While \cref{FIG:1} depicts dense skip connections for completeness, our implementation explicitly uses connections only between the encoder-decoder pairs and their immediate preceding and subsequent levels to balance computational complexity and feature aggregation. Mathematically, these skip connections are defined as:
\begin{equation}
Y_{D}^j = \mathcal{S_F} \left( \left[ \mathcal{K}(Y_{E}^j), \mathcal{P}(Y_{D}^{j+1}), \mathcal{Q}(Y_{D}^{j-1}) \right] \right),
\end{equation}
where $Y_{D}^j$ and $Y_{E}^j$ represent the feature maps at the $j$-th level of the decoder and encoder, $\mathcal{P}(\cdot)$ and $\mathcal{Q}(\cdot)$ perform upsampling and downsampling, and $\mathcal{K}(\cdot)$ represents convolution.

To further clarify, each decoder level aggregates information from the encoder at the same level, the decoder level above ($j+1$), and the decoder level below ($j-1$). This design efficiently incorporates both local and contextual information while minimizing redundant feature flow.

The feature map stack of $Y_{D}^j$ is represented as:
\begin{equation}
Y_{D}^j= \begin{cases}
Y_{E}^j = \text{ReLU}\left(\text{BN}\left(\text{Conv}\left(Y_{E}^{j-1}; \theta_j\right)\right)\right), & j=M, \\
\mathcal{S_F}\left([\underbrace{\mathcal{K}(\mathcal{P}(Y_{E}^l))_{l=1}^{j-1}, \mathcal{K}(Y_{E}^j)}_{\text{Scales } 1^{st} \sim j^{th}}, \underbrace{\mathcal{K}(\mathcal{Q}(Y_{D}^l))_{l=j+1}^M}_{\text{Scales: }(j+1)^{th} \sim M^{th}}]\right), & j=1, \cdots, M-1,
\end{cases}
\end{equation}
where $\mathcal{K}(\cdot)$, $\mathcal{P}(\cdot)$, and $\mathcal{Q}(\cdot)$ denote convolution, upsampling, and downsampling operations, respectively. $\mathcal{F}(\cdot)$ performs feature aggregation, including convolution followed by Batch Normalization (BN) and ReLU activation.

This setup enhances full-scale processing capabilities and reduces the excessive channel multiplication seen in traditional U-Net architectures. The parameter count for decoder stage \(j\) is given by:
\begin{equation}
N_{De}^j = D_S \times D_S \times \left[\left(\sum_{k=1}^j c\left(Y_{E}^k\right) + \sum_{k=j+1}^M c\left(Y_{D}^k\right)\right) \times 64 + c\left(Y_{D}^j\right)^2\right],
\end{equation}
where \(D_S\) is the convolution kernel size and \(c(\cdot)\) indicates the channel count at each layer.

By integrating full-scale skip connections between critical levels, FH-Seg maintains a balance between capturing hierarchical context and preserving computational efficiency, as shown schematically in \cref{FIG:1}.

\begin{figure}[!t]
\centering
\includegraphics[width=\columnwidth, height=0.45\columnwidth]{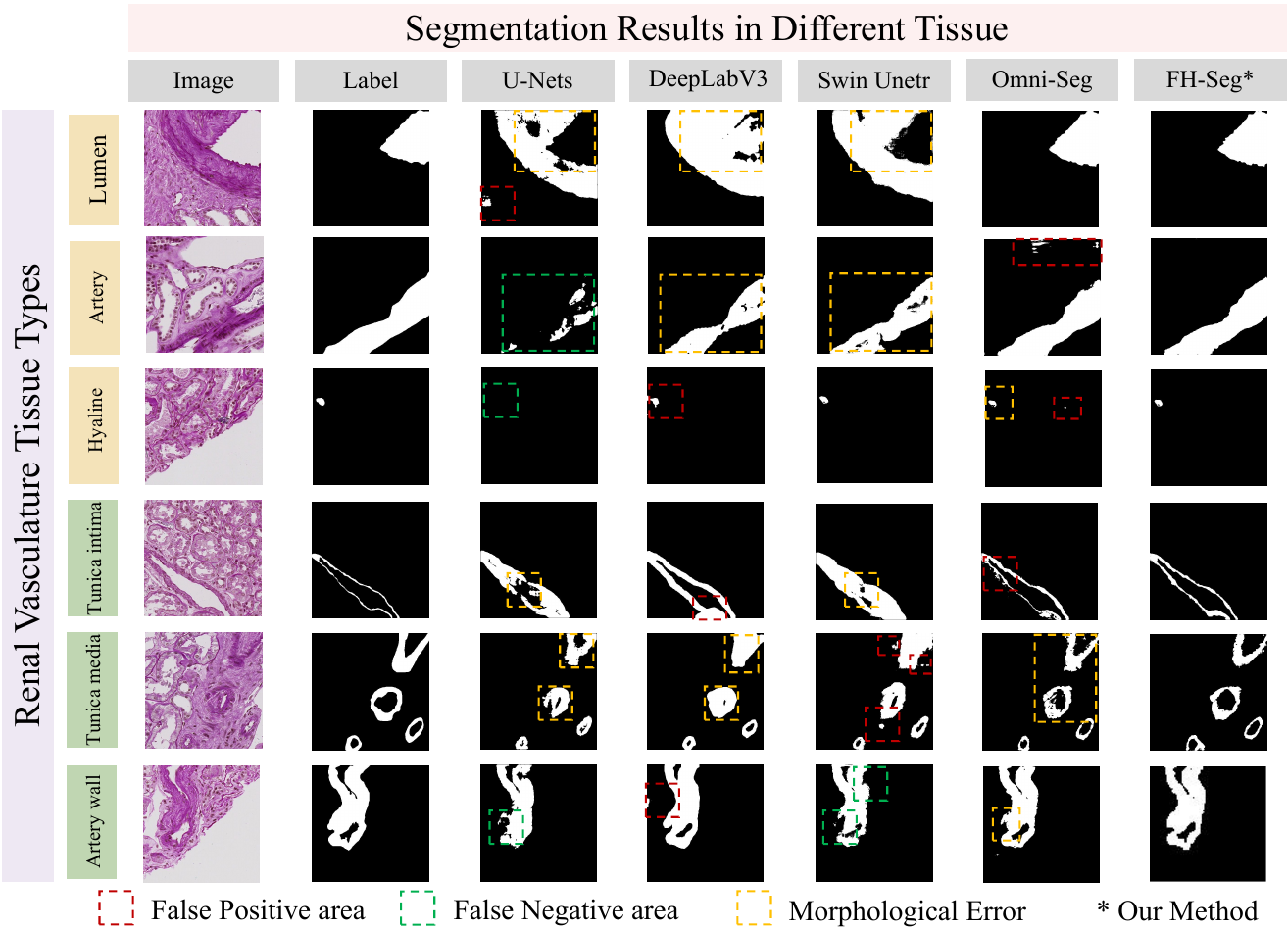}
\caption{Qualitative Comparison of Various Approaches on the LRV Dataset.}
\label{fig:Qualitative Comparison}
\end{figure}

\subsection{Learnable Hierarchical Soft Attention Gates}
Incomplete tissue segmentation can cause inconsistencies in mask labels for the same category, leading to errors in predictions. To address this, we introduce a learnable hierarchical soft-attention gate (HSA) strategy that adaptively reduces noise interference at each scale and promotes core information learning. The attention mechanism computes weights as follows:
\begin{align}
r_{att}^m &= \varphi^T\left(\sigma_1\left(M_x^T y_i^m + M_y^T h_i + c_y\right)\right) + c_\varphi, \\
\beta_i^m &= \sigma_2\left(r_{att}^m\left(y_i^m, h_i ; \Theta_{att}\right)\right),
\end{align}
where $r_{att}^m$ represents the soft attention weights, $\varphi$ is a linear transformation matrix, $\sigma_1$ and $\sigma_2$ are activation functions (ReLU and sigmoid, respectively), $M_x$ and $M_y$ are transformation matrices for the input feature map and gating signal, and $c_y$, $c_\varphi$ are bias terms.

The calculated attention weights $\beta_i^m$ are applied to feature maps via element-wise multiplication, ensuring that critical features are enhanced while irrelevant or noisy features are suppressed:
\begin{equation}
Y_{AD}^j = \mathcal{S}_F \left( \left[ \beta_i^m \odot \mathcal{K}(Y_{E}^j), \beta_i^m \odot \mathcal{U}(Y_{AD}^{j+1}), \beta_i^m \odot Y_{AD}^{j-1} \right] \right),
\end{equation}
where $\odot$ represents element-wise multiplication, $\mathcal{K}(\cdot)$ denotes convolution, and $\mathcal{U}(\cdot)$ represents upsampling.

\textbf{Resampler.}  
The Resampler ensures feature maps at different scales are compatible before calculating attention weights. It normalizes spatial resolutions and channel dimensions of input features, enabling effective gating. Mathematically, the Resampler performs the following transformations:
\begin{equation}
\tilde{y}_i^m = \mathcal{R}(y_i^m) = \text{Conv}\left(\text{Upsample}(y_i^m)\right),
\end{equation}
where $\mathcal{R}(\cdot)$ represents the resampling operation, which first upsamples the feature map to the target resolution and then applies a convolution to align the channel dimensions. This ensures seamless integration of feature maps from different scales within the HSA mechanism.

\textbf{Backpropagation Adjustment.}  
The gradients of the gated output are adjusted during backpropagation as follows:
\begin{equation}
\frac{\partial(\hat{y}_i^m)}{\partial(\Psi^{m-1})} = \beta_i^m \frac{\partial(f(y_i^{m-1} ; \Psi^{m-1}))}{\partial(\Psi^{m-1})} + \frac{\partial(\beta_i^m)}{\partial(\Psi^{m-1})} y_i^m,
\end{equation}
where $\hat{y}_i^m$ is the corrected feature map, $\beta_i^m$ is the attention coefficient, $y_i^m$ is the input feature map, and $\Psi^{m-1}$ represents the parameters of the previous layer. The first term scales the gradient by the attention coefficient, while the second term propagates the gradient adjustment from $\beta_i^m$.

\section{DATA AND EXPERIMENTS}
Due to the scarcity of labeled renal vascular pathological images and the inherent limitations of renal scintigraphy, we developed a comprehensive dataset of 16,214 annotated images. These images were segmented from high-resolution whole slide images (WSIs) of renal vascular pathology tissue, sourced from Vanderbilt University Medical Center (VUMC). Our dataset uses tissue obtained from human subjects, specifically 10 normal adults and 12 aging adults (over 65 years old), encompassing approximately 5,600 renal arteries. Unlike previously annotated WSIs introduced in other studies, our dataset stands out for its meticulous annotation of different arterial components. Expert annotators from VUMC manually identified and labeled various structures, including the lumen, tunica intima, tunica media, artery, artery wall, and hyalinosis lesion. This detailed component-based annotation makes our dataset the first to provide such fine-grained, comprehensive labels for renal vascular pathological images.

\textbf{Label Description.} The terms ``artery,'' ``artery wall,'' and ``hyaline'' refer to distinct anatomical and pathological features in our dataset. The ``artery'' encompasses the entire arterial structure, including the lumen, tunica intima, tunica media, and external elastic regions. The ``artery wall'' specifically refers to the combined tunica media and tunica intima layers, excluding the lumen. The ``hyaline'' class refers to regions of hyalinization observed within the arterial or arteriolar wall, often characterized by proteinaceous deposition. The ``lumen'' denotes the hollow interior of the artery through which blood flows. The ``tunica intima'', consisting of endothelial cells and the internal elastic lamina, forms the innermost layer, while the ``tunica media'' represents the middle layer, composed of smooth muscle cells and elastic fibers, and lies between the internal and external elastic laminae.

\begin{figure}[t] 
\centering
\begin{minipage}[t]{0.75\textwidth}
    \begin{algorithm}[H]
    \caption{Main algorithm of FH-Seg}
    \label{alg:FH-Seg}
    \begin{algorithmic}[1]
    \REQUIRE Input image $x$, Residual U-Net encoder $f_e$, decoder $f_d$, full-scale skip connections $\mathcal{S}_F$, hierarchical soft-attention gates $\mathcal{A}_{\text{HSA}}$, number of encoder layers $L$, convolution kernel size $K$, number of channels $C$
    \STATE \textbf{Initialization}: Initialize weights of encoder $f_e$ and decoder $f_d$
    \FOR{each training iteration}
        \STATE $encoder\_features = []$
        \STATE $x_e = x$
        \FOR{$i = 1$ to $L$}
            \STATE $x_e = \text{Conv}(x_e, \text{filters}=C \cdot 2^i, \text{kernel\_size}=K)$
            \STATE $x_e = \text{BatchNorm}(x_e)$
            \STATE $x_e = \text{ReLU}(x_e)$
            \STATE Append $x_e$ to $encoder\_features$
            \STATE $x_e = \text{Downsample}(x_e)$
        \ENDFOR
        \STATE $x_d = encoder\_features[-1]$
        \FOR{$j = L$ to $1$}
            \STATE $x_d = \text{Upsample}(x_d)$
            \STATE $skip\_feature = encoder\_features[j-1]$
            \STATE $x_d = \text{Concatenate}(x_d, \mathcal{S}_F(\text{skip\_feature}))$
            \STATE $x_d = \text{Conv}(x_d, \text{filters}=C \cdot 2^j, \text{kernel\_size}=K)$
            \STATE $x_d = \text{BatchNorm}(x_d)$
            \STATE $x_d = \text{ReLU}(x_d)$
        \ENDFOR
        \FOR{each feature map $Y_{D}^j$ in $x_d$}
            \STATE $attention\_weights = \mathcal{A}_{\text{HSA}}(Y_{D}^j)$
            \STATE $Y_{D}^j = Y_{D}^j \cdot attention\_weights$
        \ENDFOR
        \STATE $loss = \text{CalculateLoss}(x_d, ground\_truth)$
        \STATE $\text{Update}(f_e, f_d, loss)$
    \ENDFOR
    \end{algorithmic}
    \end{algorithm}
\end{minipage}
\end{figure}

\textbf{Labeling Methodology.} The dataset adopts a multiclass labeling approach, where each pixel is assigned to a single class, avoiding overlaps or partial labels. The ``artery" label represents the entire arterial structure and is not used simultaneously with its subcomponents (e.g., tunica intima, tunica media). Instead, these subcomponents are labeled individually to enable fine-grained segmentation. This design ensures clear separation between labels and compatibility with U-Net-based models, avoiding confusion from hierarchical annotations.

Each segmented image was derived from high-resolution WSIs consisting of several hundred thousand pixels, which were partitioned into 2048 × 2048 patches. The dataset was then randomly divided into 9,725 images for training, 3,240 for validation, and 3,249 for testing. To meet experimental requirements, further augmentation was performed by cropping to 512 × 512 pixels and applying random flips and color perturbation to enhance robustness. For more details, see \cref{tab:dataset_partition_summary} and \cref{FIG:LRV}.

\section{RESULTS}
\subsection{Comparison with State-of-the-Art Methods}
\begin{figure}[t]
    \centering
    \noindent 
    \begin{minipage}[t]{0.63\textwidth}
        \vspace*{1pt}
        \captionof{table}{Performance on Non-Hollow Structures}
        \setlength{\tabcolsep}{2pt} 
        \renewcommand{\arraystretch}{1.14} 
        \resizebox{\textwidth}{!}{%
        \begin{tabular}{c|ccc|ccc|ccc}
        \hline
        \multirow{2}{*}{Method} & \multicolumn{3}{c|}{Lumen} & \multicolumn{3}{c|}{Artery} & \multicolumn{3}{c}{Hyaline} \\
        \cline{2-10}
         & Dice & F1 & Average & Dice & F1 & Average & Dice & F1 & Average \\
        \hline
        U-Nets \cite{ronneberger2015u} & 44.58 & 52.08 & 48.33 & 62.70 & 65.80 & 64.25 & 51.47 & 63.80 & 57.64\\
        DeepLabV3 \cite{lutnick2019integrated} & 46.68 & 53.35 & 50.02 & 71.47 & 73.53 & 72.50 & 52.21 & 67.14 & 59.68 \\
        Swin Unetr\cite{hatamizadeh2021swin} & 51.74 & 52.64 & 52.19 & 67.02 & 69.22 & 68.12 & 52.71 & 62.23 & 57.47 \\
        Omni-Seg \cite{Omni-seg} & 75.54 & 76.75 & 76.15 & 72.58 & 73.70 & 73.14 & 54.55 & 68.05 & 61.30 \\
        \textbf{FH-Seg} & \textbf{77.11} & \textbf{78.15} & \textbf{78.63} & \textbf{74.75} & \textbf{75.79} & \textbf{75.27} & \textbf{55.78} & \textbf{69.44} & \textbf{62.61} \\
        \hline
        \end{tabular}
        }
        \label{tab:simple_structures}
    \end{minipage}%
    \hspace{5pt} 
    \begin{minipage}[t]{0.35\textwidth}
        \vspace*{0pt}
        \includegraphics[width=\textwidth, height=0.115\textheight]{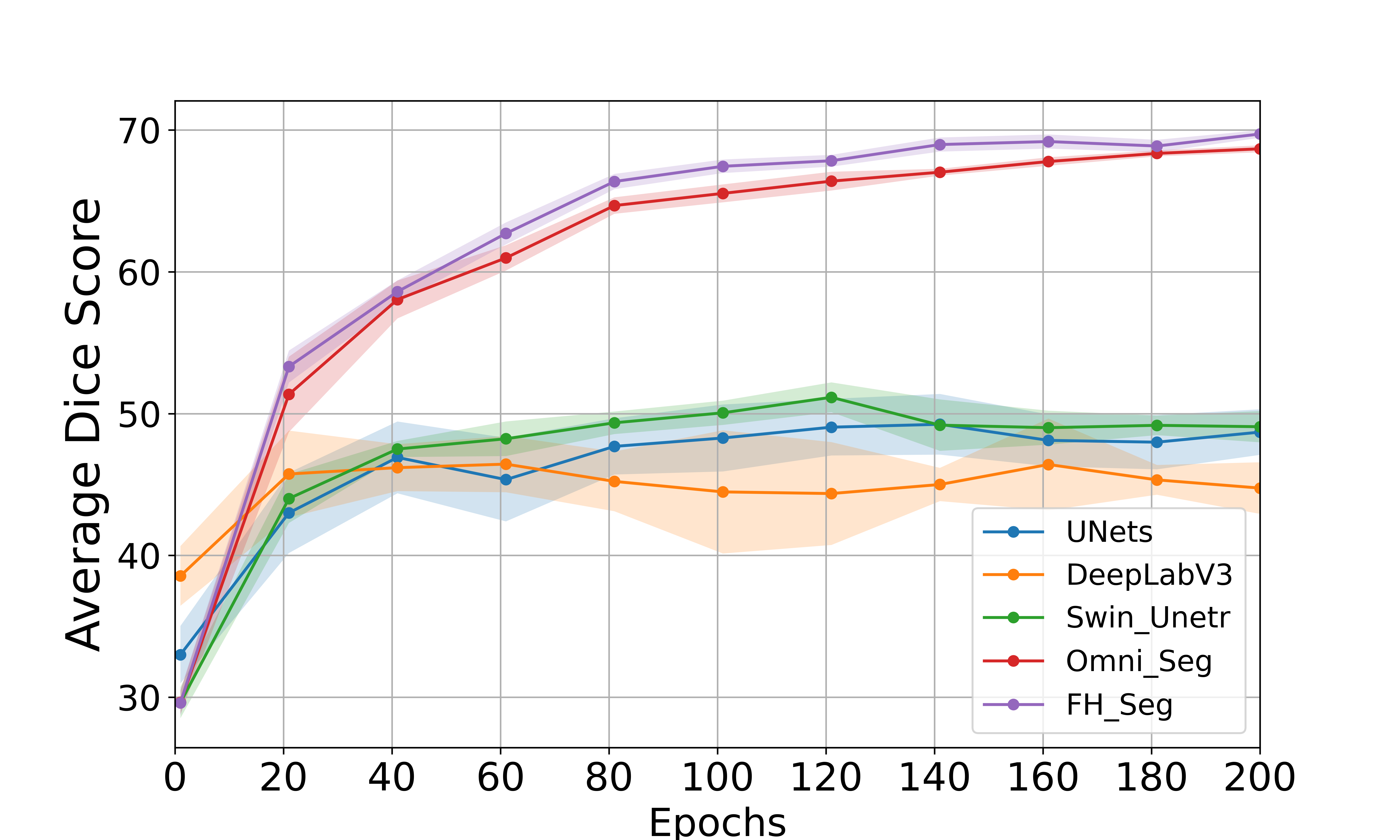}
        \caption{Non-Hollow Comparison.}
        \label{fig:non-hollow Dice mean}
    \end{minipage}
\end{figure}

\begin{figure}[t]
    \centering
    \begin{minipage}[t]{0.63\textwidth}  
        \vspace*{1pt}
        \captionof{table}{Performance on Complex Hollow Structures}
        \setlength{\tabcolsep}{2pt}
        \renewcommand{\arraystretch}{1.14}
        \resizebox{\textwidth}{!}{%
        \begin{tabular}{c|ccc|ccc|ccc}
        \hline
        \multirow{2}{*}{Method} & \multicolumn{3}{c|}{Tunica intima} & \multicolumn{3}{c|}{Tunica media} & \multicolumn{3}{c}{Artery wall} \\
        \cline{2-10}
         & Dice & F1 & Average & Dice & F1 & Average & Dice & F1 & Average \\
        \hline
        U-Nets \cite{jayapandian2021development} & 68.44 & 73.99 & 71.22 & 64.87 & 66.34 & 65.61 & 72.02 & 73.25 & 72.64 \\
        DeepLabV3 \cite{lutnick2019integrated} & 70.80 & 74.97 & 72.89 & 62.57 & 64.04 & 63.31 & 77.11 & 78.96 & 78.04 \\
        Swin Unetr\cite{hatamizadeh2021swin} & 65.84 & 67.27 & 66.56 & 61.08 & 61.84 & 61.46 & 71.13 & 72.45 & 71.79 \\
        Omni-Seg \cite{Omni-seg} & 71.56 & 72.40 & 72.48 & 68.56 & 70.93 & 69.75 & 76.23 & 78.34 & 77.29 \\
        \textbf{FH-Seg} & \textbf{73.26} & \textbf{75.75} & \textbf{74.51} & \textbf{71.23} & \textbf{73.06} & \textbf{72.15} & \textbf{78.26} & \textbf{80.48} & \textbf{79.37} \\
        \hline
        \end{tabular}
        }
        \label{tab:complex_structures}
    \end{minipage}%
    \hspace{5pt}  
    \begin{minipage}[t]{0.35\textwidth}  
        \vspace*{0pt}
        \includegraphics[width=\textwidth, height=0.115\textheight]{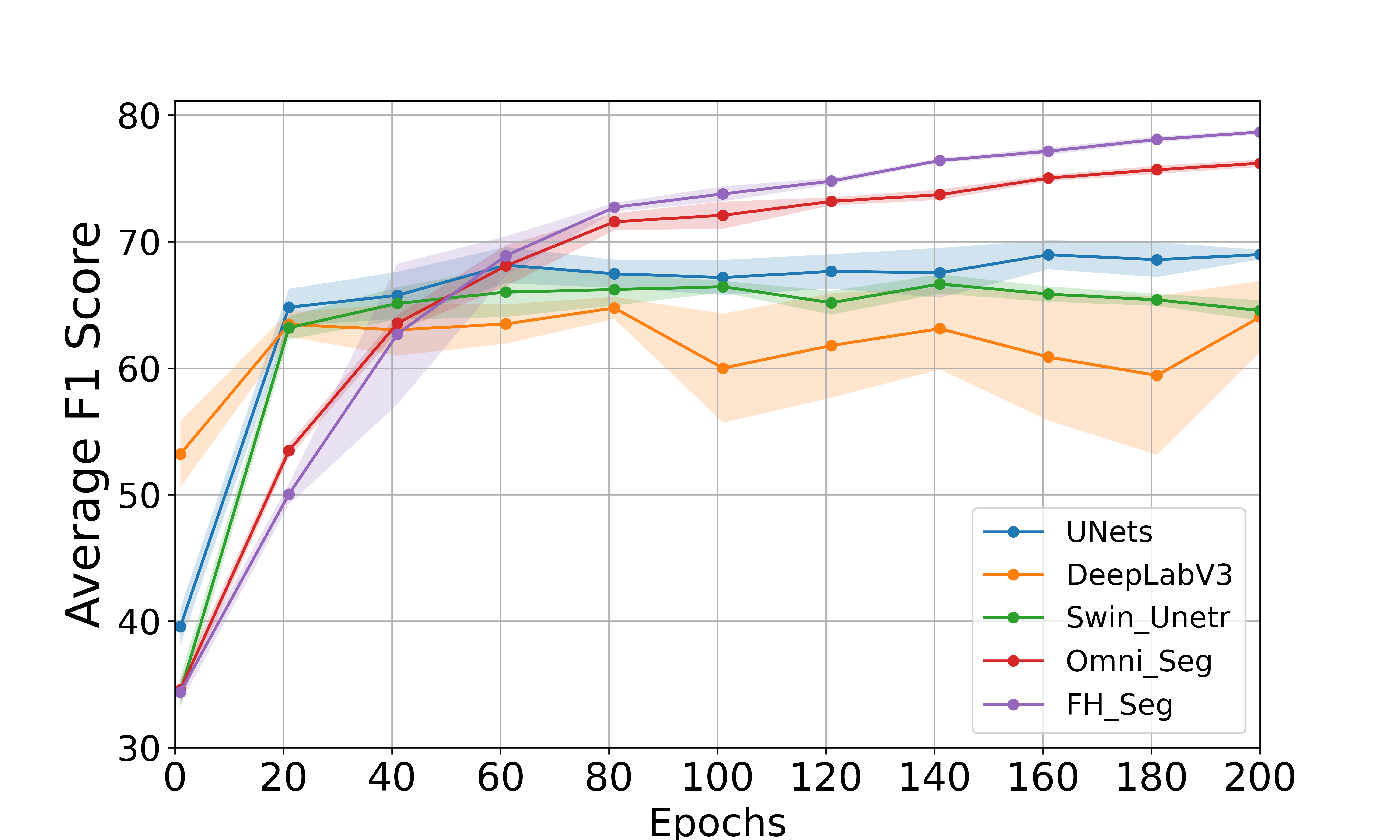}
        \caption{Hollow Comparison.}
        \label{fig:hollow F1 mean}
    \end{minipage}
\end{figure}

FH-Seg achieves superior performance on both hollow and non-hollow structures, as shown in Tables \ref{tab:complex_structures} and \ref{tab:simple_structures}. For instance, FH-Seg achieves a Dice score of 78.26\% on the artery wall, outperforming Omni-Seg by 2.03\%, and 77.11\% on the lumen, further improving upon Omni-Seg's 75.54\%. This improvement demonstrates the effectiveness of our framework in addressing challenges such as irregular lumen shapes, blurred boundaries, and noise from histological artifacts in renal pathological images. In comparison, U-Net achieves a Dice score of only 44.58\% on the lumen. This is likely due to the presence of irregular lumen shapes, blurred boundaries, and noise introduced by histological artifacts, which simple models like U-Net struggle to handle effectively. 

\subsection{Qualitative Analysis}
As shown in Figure \ref{fig:Qualitative Comparison}, FH-Seg significantly reduces false positives, false negatives, and morphological errors compared to other methods. False positives in U-Net and DeepLabV3 are largely caused by their inability to suppress noise, leading to over-segmentation in regions with similar textures. Swin Unetr and Omni-Seg, on the other hand, often miss small or irregularly shaped structures like hyaline lesions, resulting in false negatives. Morphological errors, such as fragmented or incomplete boundaries in the tunica intima and artery wall, are also more frequent in these methods due to limited feature integration.

FH-Seg addresses these challenges through full-scale skip connections, which integrate global and local contextual information, and hierarchical soft attention gates, which suppress irrelevant features while focusing on core vascular structures. This enables FH-Seg to achieve smoother boundaries, better capture small lesions, and minimize segmentation errors across all tissue types, demonstrating its robustness for renal vascular pathology segmentation.

\begin{figure}[H]
    \centering
    \begin{minipage}[t]{0.57\textwidth}
        \vspace*{0pt}
        \captionof{table}{Effects of proposed components}
        \resizebox{\textwidth}{!}{%
        \renewcommand{\arraystretch}{1.1}
        \begin{tabular}{c|ccc|ccc|ccc}
        \hline
        \multirow{2}{*}{Method} & \multicolumn{3}{c|}{Lumen} & \multicolumn{3}{c|}{Tunica intima} & \multicolumn{3}{c}{Tunica media} \\
        \cline{2-10}
         & Dice & F1 & Average & Dice & F1 & Average & Dice & F1 & Average \\
        \hline
        Base & 75.54 & 76.75 & 76.15 & 71.56 & 72.40 & 71.98 & 68.56 & 70.93 & 69.75 \\
        w/HSA & 76.22 & 77.25 & 77.04 & 72.41 & 74.08 & 73.25 & 69.89 & 72.49 & 71.19 \\
        w/FS & 76.62 & 77.55 & 77.59 & 72.93 & 74.92 & 73.93 & 70.11 & 72.99 & 71.55 \\
        Total & 77.11 & 78.15 & 78.63 & 73.26 & 75.75 & 74.51 & 71.23 & 73.06 & 72.15 \\
        \hline
        \end{tabular}
        }
        \resizebox{\textwidth}{!}{%
        \begin{tabular}{c|ccc|ccc|ccc}
        \hline
        \multirow{2}{*}{Method} & \multicolumn{3}{c|}{Artery} & \multicolumn{3}{c|}{Artery wall} & \multicolumn{3}{c}{Hyaline} \\
        \cline{2-10}
         & Dice & F1 & Average & Dice & F1 & Average & Dice & F1 & Average \\
        \hline
        Base & 72.58 & 73.70 & 73.14 & 76.23 & 78.34 & 77.29 & 54.55 & 68.05 & 61.30 \\
        w/HSA & 73.66 & 74.74 & 74.20 & 77.44 & 79.41 & 78.43 & 55.16 & 68.24 & 61.45 \\
        w/FS & 74.17 & 74.85 & 74.51 & 77.79 & 79.89 & 78.84 & 55.37 & 68.74 & 61.96 \\
        Total & 74.75 & 75.79 & 75.27 & 78.26 & 80.48 & 79.37 & 55.78 & 69.44 & 62.61 \\
        \hline
        \end{tabular}}
        \label{tab:Ablation Studies}
    \end{minipage}%
    \begin{minipage}[t]{0.35\textwidth}
        \vspace*{0pt}
        \includegraphics[width=\textwidth, height=0.135\textheight]{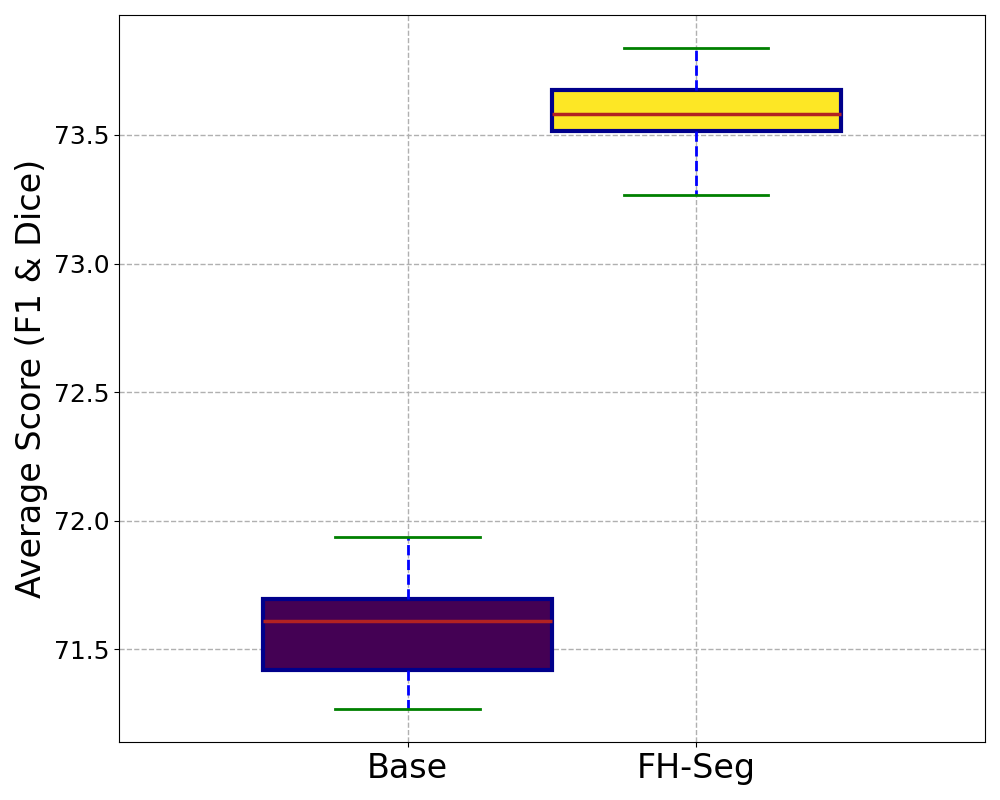}
        \caption{Bbox plot}
        \label{fig:bbox plot}
    \end{minipage}
\end{figure}

\subsection{Ablation Studies}
Table \ref{tab:Ablation Studies} shows the impact of individual components in the FH-Seg framework. The addition of hierarchical soft attention gates (HSA) improves average Dice scores by 1.1\%, while full-scale skip connections (FS) further enhance accuracy by 1.3\%. The combined architecture achieves a total improvement of 2.4\%, demonstrating the complementary benefits of these components.

\section{CONCLUSION}
In this paper, we present the FH-Seg framework, a Full-Scale Hierarchical Learning approach designed for comprehensive segmentation of the renal vasculature. FH-Seg introduces HSA after full-scale skip connections to efficiently fuse detailed anatomical information and contextual semantics across feature maps, thereby enhancing the focus on critical full-scale vascular features. To advance renal pathology segmentation research, we are also developing the LRV dataset that covers various complex parts of the renal vasculature. The FH-Seg achieves SOTA performance on the LRV dataset, with 2.67\% and 2.13\% improvements compared to Omni-Seg in Dice and F1 metrics, respectively.
\section{NEW OR BREAKTHROUGH WORK TO BE PRESENTED}
We introduce FH-Seg, a full-scale hierarchical learning framework designed for the comprehensive segmentation of the renal vasculature, and present the Large Renal Vasculature (LRV) Dataset to support advanced research in renal pathology segmentation.


\section*{Acknowledgements}
This research was supported by NIH R01DK135597(Huo), DoD HT9425-23-1-0003(HCY), NIH NIDDK DK56942 (ABF). This work was also supported by Vanderbilt Seed Success Grant, Vanderbilt Discovery Grant, and VISE Seed Grant. This project was supported by The Leona M. and Harry B. Helmsley Charitable Trust grant G-1903-03793 and G-2103-05128. This research was also supported by NIH grants R01EB033385, R01DK132338, REB017230, R01MH125931, and NSF 2040462. We extend gratitude to NVIDIA for their support by means of the NVIDIA hardware grant. This works was also supported by NSF NAIRR Pilot Award NAIRR240055.


\bibliography{report} 

\begin{thebibliography}{10}

\bibitem{gomes2021building}
Gomes, J., Kong, J., Kurc, T., Melo, A.~C., Ferreira, R., Saltz, J.~H., and Teodoro, G., ``Building robust pathology image analyses with uncertainty quantification,'' {\em Computer Methods and Programs in Biomedicine}~{\bf 208},  106291 (2021).

\bibitem{marti2021digital}
Marti-Aguado, D., Rodr{\'\i}guez-Ortega, A., Mestre-Alagarda, C., Bauza, M., Valero-P{\'e}rez, E., Alfaro-Cervello, C., Benlloch, S., P{\'e}rez-Rojas, J., Ferr{\'a}ndez, A., Alemany-Monraval, P., et~al., ``Digital pathology: accurate technique for quantitative assessment of histological features in metabolic-associated fatty liver disease,'' {\em Alimentary Pharmacology \& Therapeutics}~{\bf 53}(1),  160--171 (2021).

\bibitem{ronneberger2015u}
Ronneberger, O., Fischer, P., and Brox, T., ``U-net: Convolutional networks for biomedical image segmentation,'' in [{\em Medical image computing and computer-assisted intervention--MICCAI 2015: 18th international conference, Munich, Germany, October 5-9, 2015, proceedings, part III 18}{\nolinebreak\hspace{0.1em}]},   234--241, Springer (2015).

\bibitem{oktay2018attention}
Oktay, O., Schlemper, J., Folgoc, L.~L., Lee, M., Heinrich, M., Misawa, K., Mori, K., McDonagh, S., Hammerla, N.~Y., Kainz, B., et~al., ``Attention u-net: Learning where to look for the pancreas,'' {\em arXiv preprint arXiv:1804.03999}  (2018).

\bibitem{huang2020unet}
Huang, H., Lin, L., Tong, R., Hu, H., Zhang, Q., Iwamoto, Y., Han, X., Chen, Y.-W., and Wu, J., ``Unet 3+: A full-scale connected unet for medical image segmentation,'' in [{\em ICASSP 2020-2020 IEEE international conference on acoustics, speech and signal processing (ICASSP)}{\nolinebreak\hspace{0.1em}]},   1055--1059, IEEE (2020).

\bibitem{jayapandian2021development}
Jayapandian, C.~P., Chen, Y., Janowczyk, A.~R., Palmer, M.~B., Cassol, C.~A., Sekulic, M., Hodgin, J.~B., Zee, J., Hewitt, S.~M., O’Toole, J., et~al., ``Development and evaluation of deep learning--based segmentation of histologic structures in the kidney cortex with multiple histologic stains,'' {\em Kidney international}~{\bf 99}(1),  86--101 (2021).

\bibitem{lutnick2019integrated}
Lutnick, B., Ginley, B., Govind, D., McGarry, S.~D., LaViolette, P.~S., Yacoub, R., Jain, S., Tomaszewski, J.~E., Jen, K.-Y., and Sarder, P., ``An integrated iterative annotation technique for easing neural network training in medical image analysis,'' {\em Nature machine intelligence}~{\bf 1}(2),  112--119 (2019).

\bibitem{hatamizadeh2021swin}
Hatamizadeh, A., Nath, V., Tang, Y., Yang, D., Roth, H.~R., and Xu, D., ``Swin unetr: Swin transformers for semantic segmentation of brain tumors in mri images,'' in [{\em International MICCAI brainlesion workshop}{\nolinebreak\hspace{0.1em}]},   272--284, Springer (2021).

\bibitem{he2020dense}
He, Y., Yang, G., Yang, J., Chen, Y., Kong, Y., Wu, J., Tang, L., Zhu, X., Dillenseger, J.-L., Shao, P., et~al., ``Dense biased networks with deep priori anatomy and hard region adaptation: Semi-supervised learning for fine renal artery segmentation,'' {\em Medical image analysis}~{\bf 63},  101722 (2020).

\bibitem{lutnick2021mo077}
Lutnick, B., Moos, K., Seshan, S.~V., Kers, J., Roelofs, J., Hellmich, M., Sciascia, S., Cicalese, P.~A., Ginley, B., Sarder, P., et~al., ``Mo077 automatic segmentation of arteries, arterioles and glomeruli in native biopsies with thrombotic microangiopathy and other vascular diseases,'' {\em Nephrology Dialysis Transplantation}~{\bf 36}(Supplement\_1),  gfab078--0013 (2021).

\bibitem{Wang2022An}
Wang, H., Huang, Z., Ye, J., Tu, C., Yang, Y., Du, S., Deng, Z., Ma, C., Niu, J., and He, J., ``An evaluation of u-net in renal structure segmentation,'' {\em ArXiv}~{\bf abs/2209.02247} (2022).

\bibitem{he2021meta}
He, Y., Yang, G., Yang, J., Ge, R., Kong, Y., Zhu, X., Zhang, S., Shao, P., Shu, H., Dillenseger, J.-L., et~al., ``Meta grayscale adaptive network for 3d integrated renal structures segmentation,'' {\em Medical image analysis}~{\bf 71},  102055 (2021).

\bibitem{xu2023hybrid}
Xu, P., Holstein-Rathlou, N.-H., S{\o}gaard, S.~B., Gundlach, C., S{\o}rensen, C.~M., Erleben, K., Sosnovtseva, O., and Darkner, S., ``A hybrid approach to full-scale reconstruction of renal arterial network,'' {\em Scientific Reports}~{\bf 13}(1),  7569 (2023).

\bibitem{mounier2002cortical}
Mounier-Vehier, C., Lions, C., Devos, P., Jaboureck, O., Willoteaux, S., Carre, A., and Beregi, J.-P., ``Cortical thickness: an early morphological marker of atherosclerotic renal disease,'' {\em Kidney international}~{\bf 61}(2),  591--598 (2002).

\bibitem{bellomo2012acute}
Bellomo, R., Kellum, J.~A., and Ronco, C., ``Acute kidney injury,'' {\em The Lancet}~{\bf 380}(9843),  756--766 (2012).

\bibitem{jimenez2006mast}
Jim{\'e}nez-Heffernan, J., Bajo, M.~A., Perna, C., del Peso, G., Larrubia, J.~R., Gamallo, C., S{\'a}nchez-Tomero, J., L{\'o}pez-Cabrera, M., and Selgas, R., ``Mast cell quantification in normal peritoneum and during peritoneal dialysis treatment,'' {\em Archives of pathology \& laboratory medicine}~{\bf 130}(8),  1188--1192 (2006).

\bibitem{Omni-seg}
Deng, R., Liu, Q., Cui, C., Yao, T., Long, J., Asad, Z., Womick, R.~M., Zhu, Z., Fogo, A.~B., Zhao, S., Yang, H., and Huo, Y., ``Omni-seg: A scale-aware dynamic network for renal pathological image segmentation,'' {\em IEEE Transactions on Biomedical Engineering}~{\bf 70}(9),  2636--2644 (2023).

\bibitem{matos2024cpp}
Matos, C. E.~F., Junior, G.~B., de~Almeida, J. D.~S., and de~Paiva, A.~C., ``Cpp-unet: Combined pyramid pooling modules in the u-net network for kidney, tumor and cyst segmentation,'' {\em IEEE Latin America Transactions}~{\bf 22}(8),  642--650 (2024).

\bibitem{cao2024rasnet}
Cao, G., Sun, Z., Wang, C., Geng, H., Fu, H., Yin, Z., and Pan, M., ``Rasnet: Renal automatic segmentation using an improved u-net with multi-scale perception and attention unit,'' {\em Pattern Recognition}~{\bf 150},  110336 (2024).

\bibitem{cai2022ma}
Cai, Y. and Wang, Y., ``Ma-unet: An improved version of unet based on multi-scale and attention mechanism for medical image segmentation,'' in [{\em Third international conference on electronics and communication; network and computer technology (ECNCT 2021)}{\nolinebreak\hspace{0.1em}]},   {\bf 12167},  205--211, SPIE (2022).

\bibitem{salvi2020karpinski}
Salvi, M., Mogetta, A., Meiburger, K.~M., Gambella, A., Molinaro, L., Barreca, A., Papotti, M., and Molinari, F., ``Karpinski score under digital investigation: a fully automated segmentation algorithm to identify vascular and stromal injury of donors’ kidneys,'' {\em Electronics}~{\bf 9}(10),  1644 (2020).

\bibitem{huo2021ai}
Huo, Y., Deng, R., Liu, Q., Fogo, A.~B., and Yang, H., ``Ai applications in renal pathology,'' {\em Kidney international}~{\bf 99}(6),  1309--1320 (2021).

\bibitem{feng2023artificial}
Feng, C. and Liu, F., ``Artificial intelligence in renal pathology: current status and future,'' {\em Biomolecules and Biomedicine}~{\bf 23}(2),  225 (2023).

\bibitem{hara2022evaluating}
Hara, S., Haneda, E., Kawakami, M., Morita, K., Nishioka, R., Zoshima, T., Kometani, M., Yoneda, T., Kawano, M., Karashima, S., et~al., ``Evaluating tubulointerstitial compartments in renal biopsy specimens using a deep learning-based approach for classifying normal and abnormal tubules,'' {\em PloS one}~{\bf 17}(7),  e0271161 (2022).

\bibitem{bueno2020glomerulosclerosis}
Bueno, G., Fernandez-Carrobles, M.~M., Gonzalez-Lopez, L., and Deniz, O., ``Glomerulosclerosis identification in whole slide images using semantic segmentation,'' {\em Computer methods and programs in biomedicine}~{\bf 184},  105273 (2020).

\bibitem{salvi2021automated}
Salvi, M., Mogetta, A., Gambella, A., Molinaro, L., Barreca, A., Papotti, M., and Molinari, F., ``Automated assessment of glomerulosclerosis and tubular atrophy using deep learning,'' {\em Computerized Medical Imaging and Graphics}~{\bf 90},  101930 (2021).

\bibitem{bouteldja2021deep}
Bouteldja, N., Klinkhammer, B.~M., B{\"u}low, R.~D., Droste, P., Otten, S.~W., Von~Stillfried, S.~F., Moellmann, J., Sheehan, S.~M., Korstanje, R., Menzel, S., et~al., ``Deep learning--based segmentation and quantification in experimental kidney histopathology,'' {\em Journal of the American Society of Nephrology}~{\bf 32}(1),  52--68 (2021).

\bibitem{azad2022smu}
Azad, R., Khosravi, N., and Merhof, D., ``Smu-net: Style matching u-net for brain tumor segmentation with missing modalities,'' in [{\em International Conference on Medical Imaging with Deep Learning}{\nolinebreak\hspace{0.1em}]},   48--62, PMLR (2022).

\bibitem{fan2020ma}
Fan, T., Wang, G., Li, Y., and Wang, H., ``Ma-net: A multi-scale attention network for liver and tumor segmentation,'' {\em IEEE Access}~{\bf 8},  179656--179665 (2020).

\bibitem{guo2021sa}
Guo, C., Szemenyei, M., Yi, Y., Wang, W., Chen, B., and Fan, C., ``Sa-unet: Spatial attention u-net for retinal vessel segmentation,'' in [{\em 2020 25th international conference on pattern recognition (ICPR)}{\nolinebreak\hspace{0.1em}]},   1236--1242, IEEE (2021).

\bibitem{azad2021deep}
Azad, R., Bozorgpour, A., Asadi-Aghbolaghi, M., Merhof, D., and Escalera, S., ``Deep frequency re-calibration u-net for medical image segmentation,'' in [{\em Proceedings of the IEEE/CVF International Conference on Computer Vision}{\nolinebreak\hspace{0.1em}]},   3274--3283 (2021).

\bibitem{Deng2023omni}
Deng, R., Liu, Q., Cui, C., Yao, T., Long, J., and Asad, Z., ``Omni-seg: A scale-aware dynamic network for renal pathological image segmentation,'' {\em IEEE Transactions on Biomedical Engineering}~{\bf 70}(9),  2636--2644 (2023).

\bibitem{zhou2018unet++}
Zhou, Z., Rahman~Siddiquee, M.~M., Tajbakhsh, N., and Liang, J., ``Unet++: A nested u-net architecture for medical image segmentation,'' in [{\em Deep Learning in Medical Image Analysis and Multimodal Learning for Clinical Decision Support: 4th International Workshop, DLMIA 2018, and 8th International Workshop, ML-CDS 2018, Held in Conjunction with MICCAI 2018, Granada, Spain, September 20, 2018, Proceedings 4}{\nolinebreak\hspace{0.1em}]},   3--11, Springer (2018).

\end{thebibliography}
\bibliographystyle{spiebib} 

\end{document}